# Edge Detection Based on Global and Local Parameters of the Image

Andrew Franklin Caetano Brustolin

Graduated in Electric Engineering from University of Uberaba

`andrewbrustolin@gmail.com`

**Abstract.** This paper presents an edge detection method based on global and local parameters of the image, which produces satisfactory results on the edge detection of complex images and has a simple structure for execution. The local and global parameters of the image are arithmetic means and standard deviations, the former acquired from a three sized window representing five pixels, the latter acquired from the entire row or column. We obtain the differences of grayscale intensities between two adjacent pixels and the sum of the modulus of these differences from the horizontal and vertical scans of the image. Using these obtained values, we calculate the local and global parameters. After the gathering of the local and global parameters, we compare each sum of the modulus of differences with its own local and global parameter. In the case of the comparison is true, the consecutive pixel to the modulus sum of differences index is marked as an edge. We present the results of the tests with grayscale images using different parameters and discuss the advantages and disadvantages of each parameter value and algorithm structure chosen on the edge processing. There is a comparison of results between this papers' detector and Canny's, where we evaluate the quality of the presented detector.

## 1 Introduction:

We can define the edges inside an image as being discontinuities of the intensities of the pixels in the image. Such discontinuities can determinate objects boundaries, change in textures, depth and luminosity of the external captured reality [6]. The importance of edge detection comes from the fact that it can represent the boundaries of different objects captured by the camera, eliminating everything else, thus greatly reducing the data and information in the image. It is important that the edges detected represent with good quality the boundaries of the objects in the pictured reality and that is suppressed the noise that could be detected as an edge, because some high level computer vision algorithms depends on the information extracted from the edge detection as first step on their execution [3].

Researchers developed and are developing edge detectors following different principles, some being more popular due to their quality on detecting the edges of an image. One of them is the Canny edge detector, which bases on a multi stage process. We can resume this process into five stages. Gaussian smoothing of the image, for noise removal. Finding the intensity gradients. Non-maximum suppression, to mark only local maxima as an edge. Double thresholding, to determine a gradient intensity range for possible edges. Edge tracking, to eliminate weak edges from strong ones [2].

The edge detector presented here bases itself on a double, one dimensional scan, of rows and columns, for the gathering of variables that represents the variation level of pixels intensities, those variables could be local, which represents five pixels, or global, which represents the entire row or column. We compare these local and global variables, for edge detection, with the modulus sum of differences from gray scale intensities of two consecutive pixels. The edges detected from the horizontal scan come from horizontal discontinuities of the pixels intensities and the vertical ones come from vertical discontinuities. We developed this paper to present a new method of edge detection, based on local and global parameters of an image, capable to detect edges on complex images with a satisfactory level of quality and simple structure for execution.

## 2   Noise Reduction:

The noise level caused during the image acquisition depends of intrinsic factors, as the camera's sensor temperature, and extrinsic factors, as the luminosity level of the area of the pictured image. During the process of edge detection, the level of noise acts in a negative way, because depending on the edge detection sensibility, this could detect a pixel affected by noise as an edge, so reducing the quality of the detector due to edges' false positives [5].

For that reason, this edge detector preprocesses the image, using the Gaussian filter, it reduces the level of noise of the image, we implement it using the OpenCV's library function GaussianBlur() [8], with 7x7 kernel size and its default standard deviation. We have chosen the 7x7-kernel size due to its good results on reducing the image noise without dislocating the possible edges from its original position [10]. Done the image preprocessing, we can execute the edge detector.

## 3   Edge Detector:

We converted the intensities values of the pixels to grayscale or worked directly with an image in grayscale for the implementation of the edge detector. We implemented the algorithm on C++, using the Codeblocks IDE and the image processing and computer vision library, OpenCV.

### 3.1   Calculation of the Modulus Sum of Differences:

After the application of the Gaussian filter, we will scan the image on the vertical and horizontal directions, on only one dimension, to calculate the intensity difference between the posterior and actual pixel. This difference is denominated in this paper as dA(r,c), where A is the matrix of pixel intensities of the image with m rows and n columns(A(m,n)), r and c are the rows and the columns indexes respectively, the differences calculation on each row and column is done in the following way for the two directions:

Horizontal scan:
dA(r,c) = A(r,c+1) − A(r,c)

Vertical scan:
dA(r,c) = A(r+1,c) − A(r,c)

After the differences calculation, we compute the sum of the modulus of these differences (smd). Below we present how to do the calculation on each direction of scan:

Horizontal scan:
smd(r,c) = |dA(r,c+1)| + |dA(r,c)|

Vertical scan:
smd(r,c) = |dA(r+1,c)| + |dA(r,c)|

### 3.2   Global Parameters:

Done the calculation of those variables on the two scan directions, to all rows and columns of the image, we compute the global arithmetic mean (mgsmd) of all the modulus sum of differences on each row or column depending on the scan direction.

Horizontal scan:
$$\text{mgsmd}(r) = \frac{\sum_{c=0}^{n-3} smd(r,c)}{n-2} \quad (3.1)$$

Vertical scan:
$$\text{mgsmd}(c) = \frac{\sum_{r=0}^{m-3} smd(r,c)}{m-2} \quad (3.2)$$

We calculate the global standard deviation (dpg) using the mgsdm on each row or column:

Horizontal scan:
$$\text{dpg}(r) = \sqrt{\frac{\sum_{c=0}^{n-3}(mgsmd(r)-smd(r,c))^2}{n-3}} \quad (3.3)$$

Vertical scan:
$$\text{dpg}(c) = \sqrt{\frac{\sum_{r=0}^{m-3}(mgsmd(c)-smd(r,c))^2}{m-3}} \quad (3.4)$$

### 3.3 Local Parameters:

After the global parameters calculations, we do again a vertical and horizontal scan, to find the local means (mlsmd) and standard deviations (dpl) from each three modulus sum of differences. Each three smd represents five pixels of the local parameters evaluation window of A(r,c), as it is shown below:

A: (r,c) (r,c+1) (r,c+2) (r,c+3) (r,c+4);  
dA(r,c) = ((r,c+1) − (r,c));  
dA(r,c+1) = ((r,c+2) − (r,c+1));  
dA(r,c+2) = ((r,c+3) − (r,c+2));  
dA(r,c+3) = ((r,c+4) − (r,c+3));  
smd(r,c) = ((|(r,c+1) − (r,c)|) + (|(r,c+2) − (r,c+1)|));  
smd(r,c+1) = ((|(r,c+2) − (r,c+1)|) + (|(r,c+3) − (r,c+2)|));  
smd(r,c+2) = ((|(r,c+3) − (r,c+2)|) + (|(r,c+4) − (r,c+3)|));

As it was shown above, the third smd of the local mean window, has an index equals to i -2, i being the fifth pixel of the local parameters evaluation window of A(r,c). Based on these three smd values, we calculate the arithmetic mean of the evaluation window:

Horizontal scan:
$$\text{mlsmd} = \frac{\sum_{c}^{c+2} smd(r,c)}{3} \quad (3.5)$$

Vertical scan:
$$\text{mlsmd} = \frac{\sum_{r}^{r+2} smd(r,c)}{3} \quad (3.6)$$

After each local mean calculation, it is evaluated the standard deviation of each window.

Horizontal scan:
$$\text{dpl(r,c)} = \sqrt{\frac{\sum_{c}^{c+2}(mlsmd-smd(r,c))^2}{3}} \quad (3.7)$$

Vertical scan:
$$\text{dpl(r,c)} = \sqrt{\frac{\sum_{r}^{r+2}(mlsmd-smd(r,c))^2}{3}} \quad (3.8)$$

After the local standard deviation evaluation, the local evaluation window advances one pixel and we calculate new local means and standard deviations. In the case of an edge is detected in that local window, the next window is going to advance four pixels (c = c + 4 or r = r + 4), the reason to this will be explained in the section IV, with the comparison between the window advancing one and three pixels when there is an edge.

### 3.4 Edge Detection Comparison inside the Evaluation Window:

During the horizontal and vertical scans, after each local mean and standard deviation evaluation, there will be a comparison between the first smd of the local evaluation window, smd(r,c), and the sum of the local mean and standard deviation(mlsmd + dpl). The value of smd(r,c) will also be compared to the rows or columns global variables (mgsmd + (thres x dpg)), the global standard deviation in this case is being multiplied by an empirically defined constant (thres).

We empirically define the thres value, higher values eliminate more edges caused by noise, but also eliminates real edges of the image, lower values permits more noise to pass as edges and gets more details of the image as edges. In section IV, we will show edges images with different values of thres for comparison.

Rows or columns that have few differences of pixel intensities (like blue skies) detects false edges on them, due to their little mgsmd value. We implemented a comparison between each mgsmd to a constant named here as thres2, and set its value very low (like 1), so if mgsmd is lesser than it, independently of the other comparisons inside the evaluation window, there will be no edge in it. The other variable is thres3, it is implemented in the case of few true edges on a row or column that has low mgsmd value not being passed as edges, so if an mlsmd value is greater than thres3 and all the other comparisons inside the evaluation window are true, there will be an edge in it. High thres2 and thres3 values will filter out true edges, so we empirically adopted with good results the thres2 value of 1 and thres3 value of 6.

The result of the comparison will define if there is an edge or not in the evaluation window, we demonstrate the comparison in pseudo code below:

```
If {(smd(r,c) > (mlsmd + dpl)) and ((smd(r,c) > (mgsmd
+ thres*dpg)) and (mgsmd > thres2) and (mlsmd > thres3)}
```

If the condition above is true, the posterior pixel to smd(r,c), which is the second pixel of the local evaluation window, is an edge and the local window advances four pixels for new measures.

After being defined the pixels that are edges on the two scans, we create a copy of the original image and it has all its pixels intensities values converted to zero, done that, we scan the image again to mark the edge pixels with the gray scale intensity value of two hundred fifty five. In the case of the same pixel is detected as an edge, either on the horizontal or vertical scan, or on both, it will be written as an edge.

### 3.5    Example:

To illustrate better what we explained, follows the example of the horizontal scan. For the vertical one, just switch the constant row for a constant column and the variable columns for variable rows.

Suppose there is an image with eleven pixels of column size (n = 11), and its corresponding pixel intensities values of the first row (r = 0):

| col index | 0 | 1 | 2 | 3 | 4 | 5 | 6 | 7 | 8 | 9 | 10 | |
|---|---|---|---|---|---|---|---|---|---|---|---|---|
| row index | | | | | | 0 | | | | | | |
| A(m,n) | 2 | 2 | 2 | 2 | 2 | 5 | 8 | 8 | 8 | 8 | 8 | pixel int. |
| dA(m,n-1) | 0 | 0 | 0 | 0 | 3 | 3 | 0 | 0 | 0 | 0 | x | |
| smd(m,n-2) | 0 | 0 | 0 | 3 | 6 | 3 | 0 | 0 | 0 | x | x | |

*Figure 01: Table with a sample of the pixels intensities of one row of the image.*

We do the global parameters calculation:

mgsmd = (0 + 0 + 0 + 3 + 6 + 3 + 0 + 0 + 0)/9 = 1.33;
dpg = 2;
thres = 1;
thres2 = 1;
thres3 = 2;

Then we calculate the local parameters for each evaluation window, and on each window, we do a comparison for a possible edge detection.

Window(r,c):

smd(r,c) = 0, smd(r,c+1) = 0, smd(r,c+2) = 0.
mlsmd(r,c) = 0, dpl(r,c) = 0.
smd(r,c) > (mlsmd(r,c) + dpl(r,c)) and smd(r,c) > (mgsmd + thres*dpg) and (mgsmd > thres2) and (mlsmd(r,c) > thres3)?
 No, evaluation window advances to c + 1.

Window(r,c+1):

smd(r,c+1) = 0, smd(r,c+2) = 0, smd(r,c+3) = 3.
mlsmd(r,c+1) = 1, dpl(r,c+1) = 1.41.
smd(r,c+1) > (mlsmd(r,c+1) + dpl(r,c+1)) and smd(r,c+1) > (mgsmd + thres*dpg) and (mgsmd > thres2) and (mlsmd(r,c+1) > thres3)?
 No, evaluation window advances to c + 2.

Window(r,c+2):

smd(r,c+2) = 0, smd(r,c+3) = 3, smd(r,c+4) = 6.
mlsmd(r,c+2) = 3, dpl(r,c+2) = 2.44.
smd(r,c+2) > (mlsmd(r,c+2) + dpl(r,c+2)) and smd(r,c+2) > (mgsmd + thres*dpg) and (mgsmd > thres2) and (mlsmd(r,c+2) > thres3)?
 No, evaluation window advances to c + 3.

Window(r,c+3):

smd(r,c+3) = 3, smd(r,c+4) = 6, smd(r,c+5) = 3.
mlsmd(r,c+3) = 4, dpl(r,c+3) = 1.41.
smd(r,c+3) > (mlsmd(r,c+3) + dpl(r,c+3)) and smd(r,c+3) > (mgsmd + thres*dpg) and (mgsmd > thres2) and (mlsmd(r,c+3) > thres3)?
 No, evaluation window advances to c + 4.

Window(r,c+4):

smd(r,c+4) = 6, smd(r,c+5) = 3, smd(r,c+6) = 0.
mlsmd(r,c+4) = 4, dpl(r,c+4) = 1.41.
smd(r,c+4) > (mlsmd(r,c+4) + dpl(r,c+4)) and smd(r,c+4) > (mgsmd + thres*dpg) and (mgsmd > thres2) and (mlsmd(r,c+4) > thres3)?
 Yes, A(r,c+5) is an edge and evaluation window advances to c + 8.

Because in the window(r,c+4) was detected an edge, the next pixel A(r,c+5), will be highlighted as an edge. Because the window advances to c+8, there will be only smd(r,c+8) left to the next window, we need at least three pixels for local parameters computation, so the scan ends there.

### 3.6 Isolated Edges Elimination:

After the filling of edges on the copy image, there will be a new scan, this one to eliminate isolated edges. If there is no edge in its eight neighboring pixel range, the scan will eliminate it. The isolated edges elimination demands more computational power, so depending of time criticality of the application, the user could not be implement it. Below follows the pseudo code for its execution on one edge of the image:

```
If {(A(r,c+1) = 0) and (A(r,c-1) = 0) and (A(r+1,c) =
0) and (A(r-1,c) = 0) and (A(r-1,c-1) = 0) and (A(r-
1,c+1) = 0) and (A(r+1,c-1) = 0) and (A(r+1,c+1) = 0)};
    A(r,c) = 0;
```

We will show the effects of isolated edges elimination, with and without it, in the section IV.

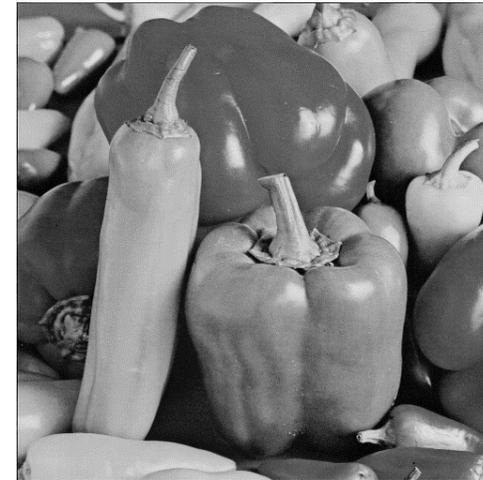
*Figure 02 [12]: Grayscale image of peppers.*

## 4 Experimental Analysis and Discussions:

In this section, we will demonstrate the results of the edge detection program on some images. There will be comparisons with different thres values, evaluation windows with four or one pixel advances and isolated edges elimination scan. The quality of this detector will be compared to the Canny's, by the execution of Lena and peppers images.

### 4.1 Analysis with Different Values of Thres:

Below is done an analysis using different values for the variable thres, we use the isolated edges elimination scan, thres2 value is 1, thres3 value is 6 and the evaluation window advances four pixels when it finds an edge.

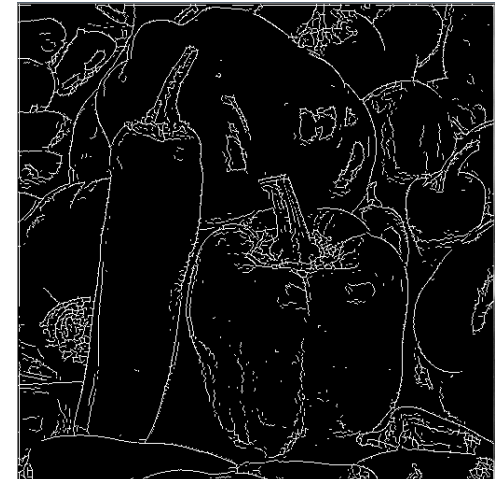
*Figure 03: Peppers: thres = 0.4, thres2 = 1, thres3 = 6.*

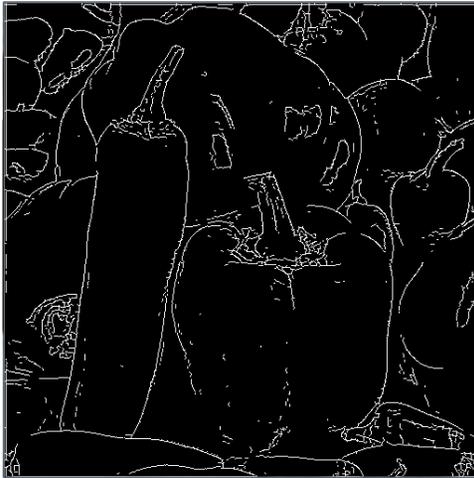

*Figure 04: Peppers: thres = 1.1, thres2 = 1, thres3 = 6.*

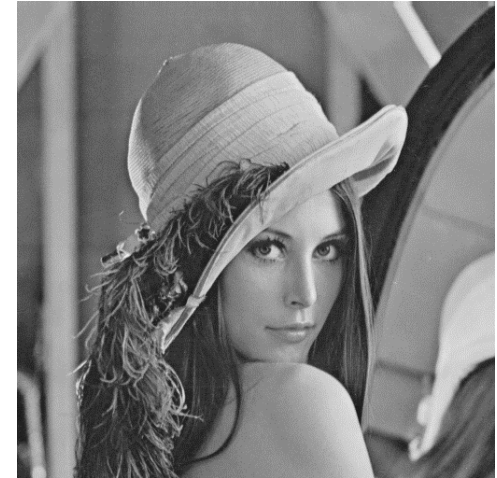

*Figure 06[12]: Lena's image*

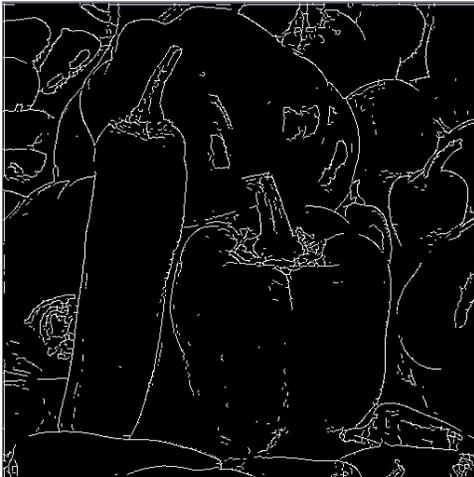

*Figure 05: Peppers: thres = 1.6, thres2 = 1, thres3 = 6.*

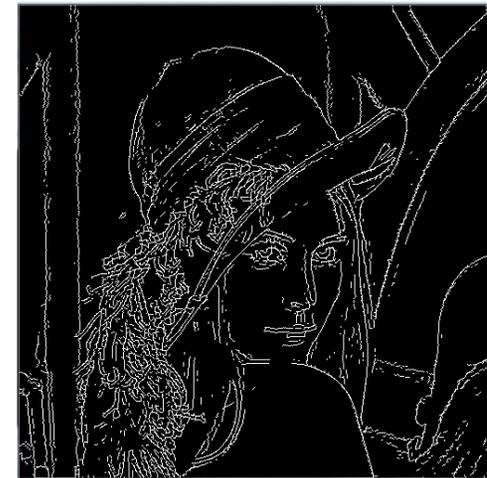

*Figure 07: Lena: thres = 0.4, thres2 = 1, thres3 = 6.*

It is visible that lower values of thres detects more details of the images and false positives, as noise, while higher values of thres detects fewer details of the image and less noise passes as edges.

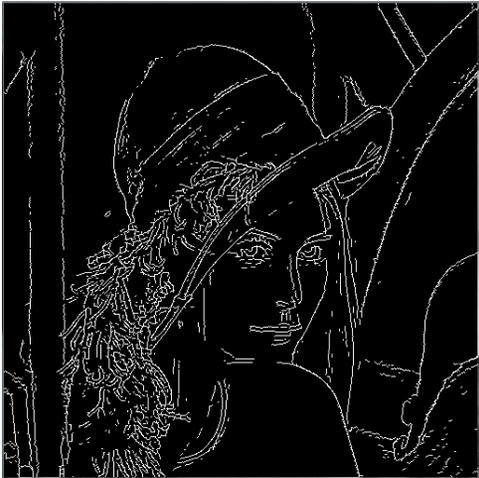

*Figure 08: Lena: thres = 0.8, thres2 = 1, thres3 = 6.*

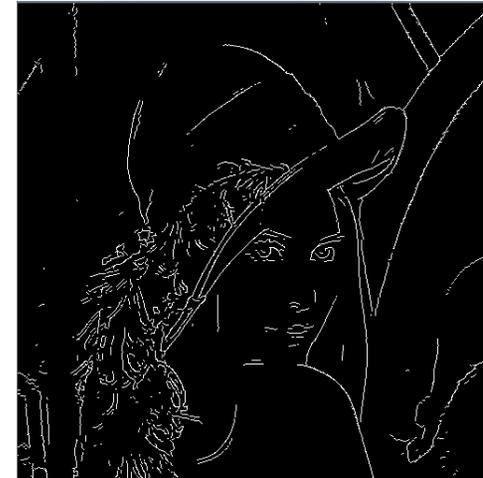

*Figure 10: Lena: thres = 2, thres2 = 1, thres3 = 6.*

### 4.2  Analysis Based on the Window Advances when there is an Edge:

Below we do an analysis based on the window advances when it finds an edge, the algorithm uses the isolated edges elimination scan, the thres value is 0.8, thres2 value is 1 and thres3 value is 6.

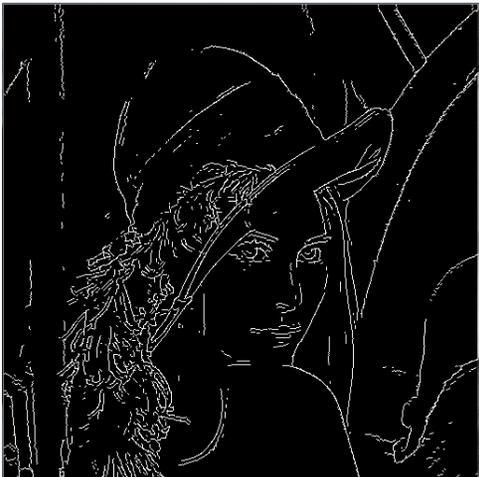

*Figure 09: Lena: thres = 1.4, thres2 = 1, thres3 = 6.*

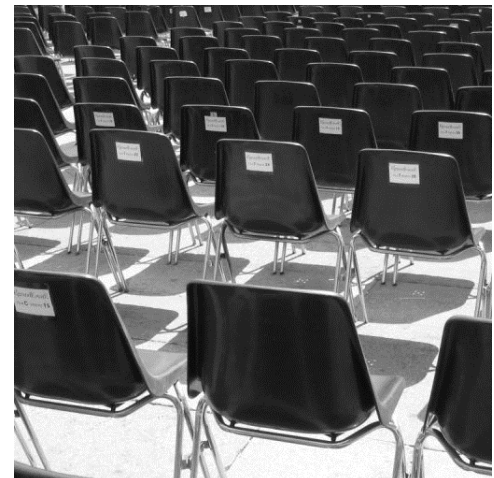

*Figure 11[4]: Image of chairs.*

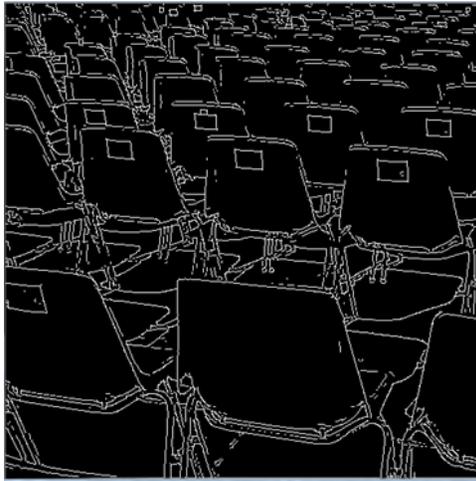

*Figure 12: thres = 0.8, thres2 = 1, thres3 = 6, evaluation window advancing four pixels when the detector finds an edge.*

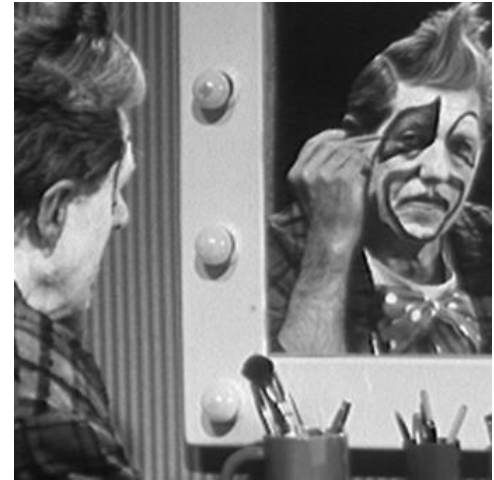

*Figure 14[7]: Image of an artist.*

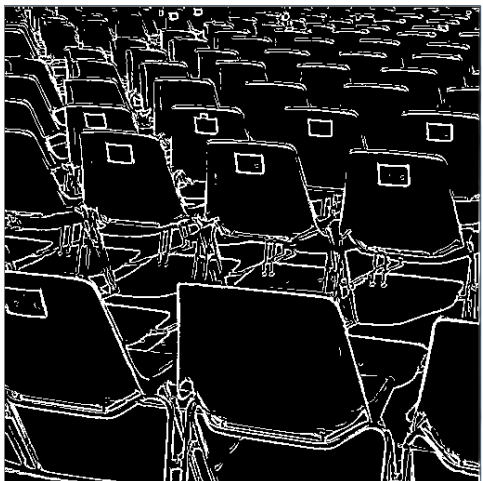

*Figure 13: thres = 0.8, thres2 = 1, thres3 = 6, evaluation window advancing one pixel when the detector finds an edge*

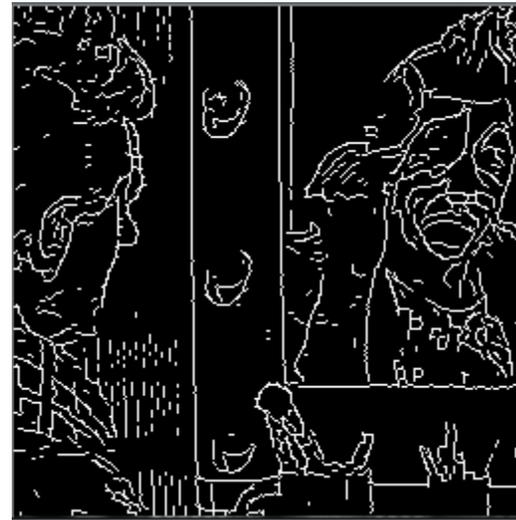

*Figure 15: thres = 0.8, thres2 = 1, thres3 = 6, evaluation window advancing four pixels when the detector finds an edge.*

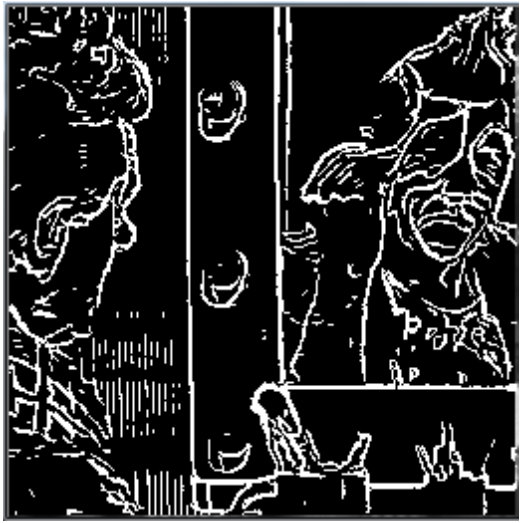

*Figure 16: thres = 0.8, thres2 = 1, thres3 = 6, evaluation window advancing one pixel when the detector finds an edge.*

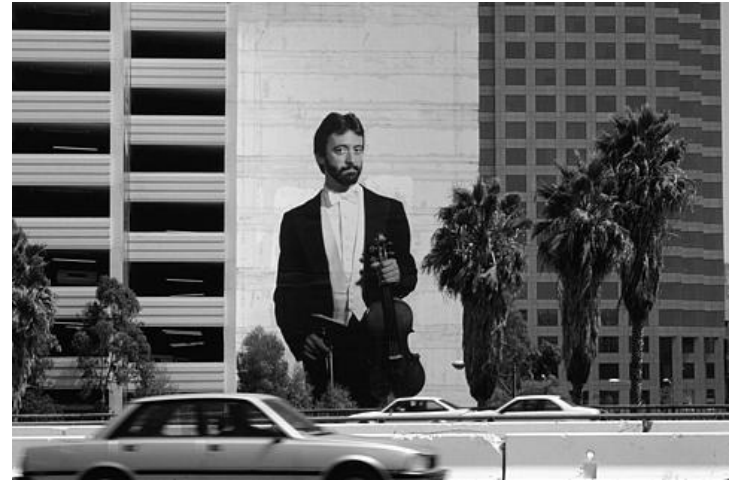

*Figure 17[1]: Man image.*

The difference of edge thickness between each pair of images, the artist and the chairs, are notable, in the images with thicker edges, there is a single pixel advance of the evaluation local window when an edge is detected, and a four-pixel advance for the thinner-edges images. Because of that, we have chosen in this algorithm, for the evaluation window to advances four pixels when it finds an edge.

Besides the fact of reducing the edge thickness, there is the argument that it can reduces the algorithm's execution time, because when more edges are found, more pixels advances are made on the scan, less pixels will be analyzed, so it will reach the end of the row or column faster, making the program faster.

The images without isolated edges elimination have little spurious isolated edge-pixels that can be eliminated, and they are, for better edge detection quality, with the isolated edge elimination scan, as in the man image, isolated edges inside the windows were eliminated, or in the house image, where the eliminated edges were on the grass.

Depending on time criticality of an application, it would be better not to apply the isolated edges elimination scan, because it takes more computational time for its execution.

### 4.3 Analysis Based on the Isolated Edges Elimination Scan usage:

Below we do an analysis based on the isolated edges elimination scan usage, the thres value is 0.8, thres2 value is 1, thres3 value is 6 and the evaluation window advances four pixels when it finds an edge.

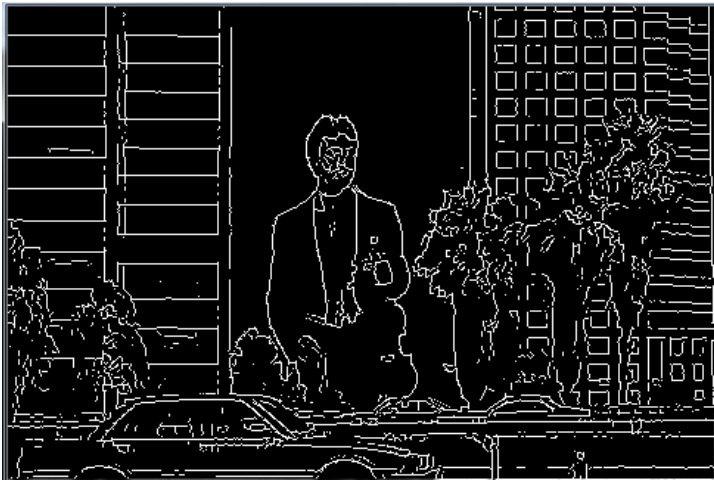

*Figure 18: thres = 0.8, thres2 = 1, thres3 = 6, without isolated edges elimination scan.*

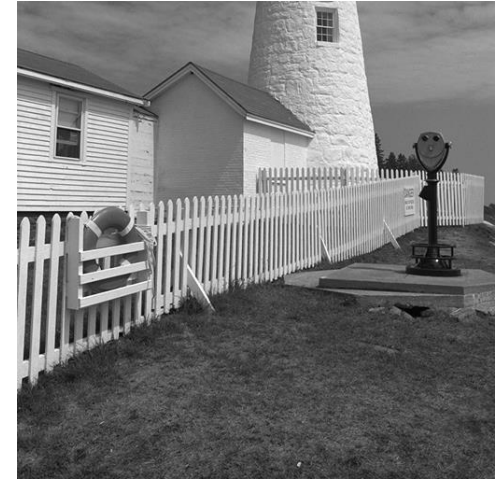

*Figure 20[1]: House image.*

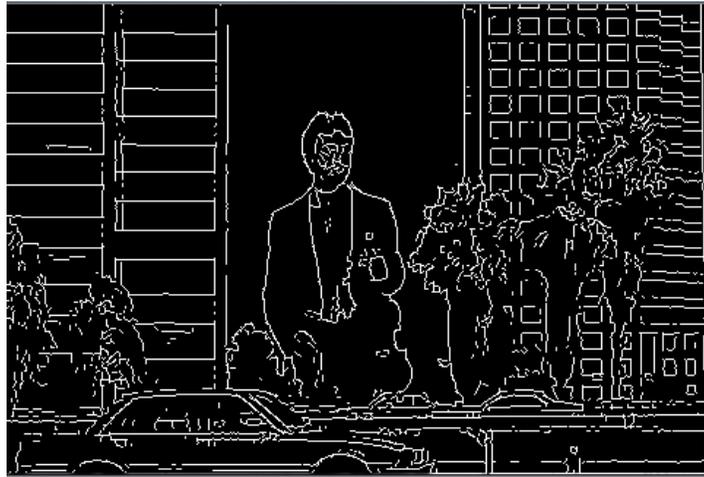

*Figure 19: thres = 0.8, thres2 = 1, thres3 = 6, with isolated edges elimination scan.*

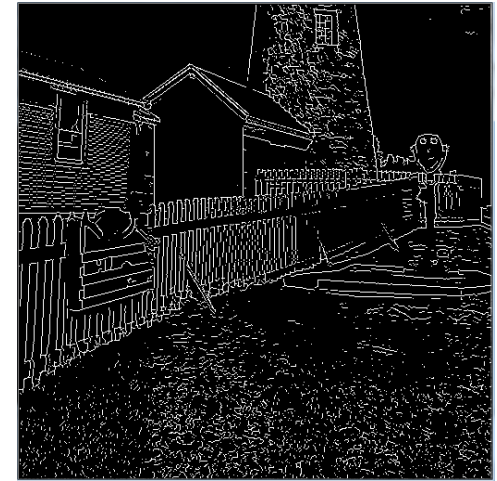

*Figure 21: thres = 0.8, thres2 = 1, thres3 = 6, without isolated edges elimination scan.*

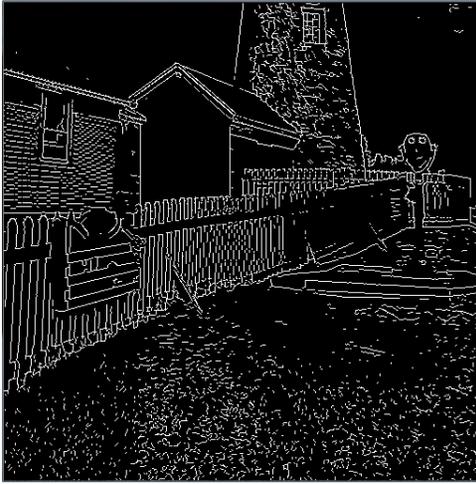

*Figure 22: thres = 0.8, thres2 = 1, thres3 = 6, with isolated edges elimination scan.*

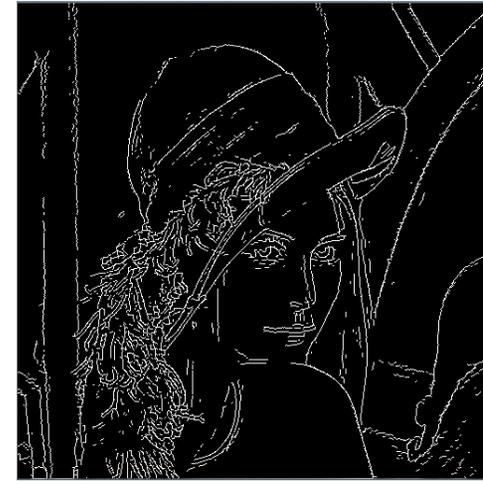

*Figure 23: Edge detection of Lena's image using this paper's method: thres = 0.8, thres2 = 1, thres3 = 6, with isolated edges elimination scan and window advancing four pixels for each edge found.*

### 4.4 Comparison with John Canny's Detector:

We are going to compare the tests results for two images with Canny's, for quality measurement of this paper edge detection method. The two images are Lena and peppers.

For Canny's edge detector implementation, we used the OpenCV function Canny() [9], its set parameters for the tests are: high threshold = 150, low threshold = 50, kernel size = 3.

This paper's detector parameters are: thres = 0.8, thres2 = 1, thres3 = 6, window advancing four pixels for each edge, with isolated edges elimination scan.

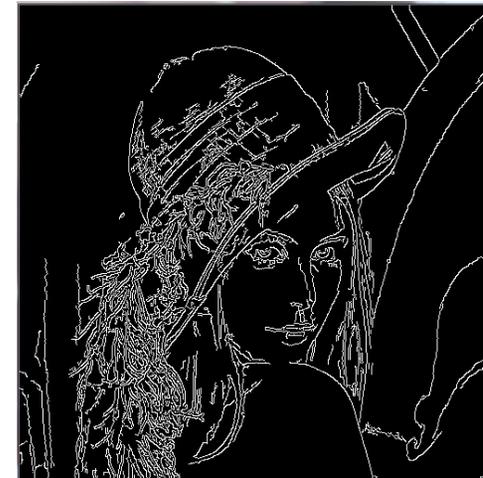

*Figure 24: Lena's image edge detection using Canny's method: high threshold = 150, low threshold = 50, kernel size = three.*

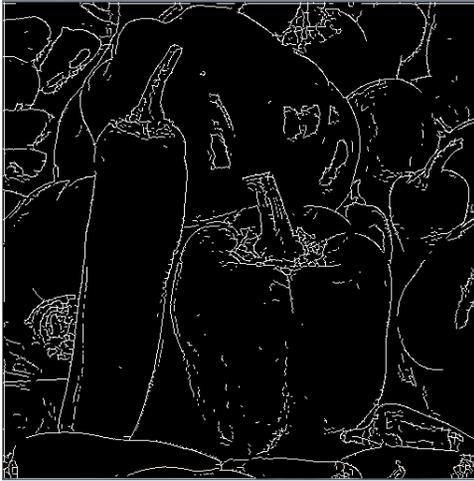

*Figure 25: Edge detection of peppers' image using this paper's method: thres = 1, thres2 = 1, thres3 = 6, with isolated edges elimination scan and window advancing four pixels for each edge found.*

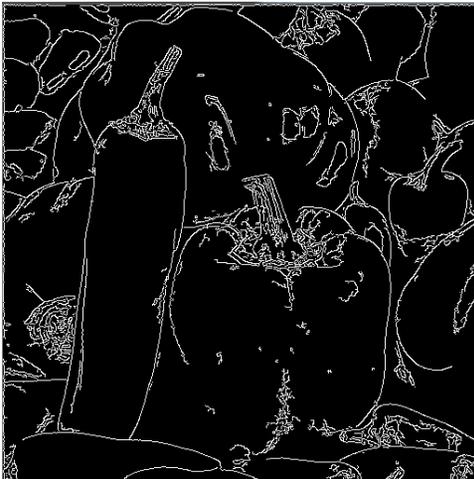

*Figure 26: Peppers' image edge detection using Canny's method: high threshold = 150, low threshold = 50, kernel size = three.*

Basing on the images above, this paper's edge detector showed itself being capable of detecting edges in a satisfactory way, the detailing level of the image presented by the edges is somewhat comparable to Canny's.

## 5  Conclusion:

The detector showed good quality in detecting edges, the edges presented themselves with thickness of one pixel, with edge detection of the main boundaries of the image, and little noise detection. On a level of algorithm's structure, one can execute the horizontal and vertical scans in parallel, turn off the isolated edges elimination scan without big consequences to the detector quality and when the detector finds an edge, the evaluation window advances four pixels, all of that resulting in a faster execution time. Based on the tests results with different parameters, the range of thres that most satisfies the condition of detecting true edges, is between 0.8 and 1.2.

## Acknowledgment:

I want to thank my family for supporting me during the time I was researching this subject and I want to thank the University of Uberaba professor Edilberto Pereira, for giving good advices on the article structure and keeping me motivated to publish it.